\documentclass{article}
\usepackage{arxiv}
\usepackage{amsmath,amsfonts}
\usepackage{algorithmic}
\usepackage{algorithm}
\usepackage{array}
\usepackage[caption=false,font=normalsize,labelfont=sf,textfont=sf]{subfig}
\usepackage{textcomp}
\usepackage{stfloats}
\usepackage{url}
\usepackage{verbatim}
\usepackage{graphicx}
\usepackage{cite}
\usepackage{hyperref}
\usepackage{academicons}
\usepackage{xcolor} 
\usepackage{multirow}
\usepackage{pifont}
\usepackage{enumitem}

\begin{document}

\title{Formulating Reinforcement Learning for Human-Robot Collaboration through Off-Policy Evaluation}


\author{Saurav Singh, 
        Rodney Sanchez, 
        Alexander Ororbia, 
        Jamison Heard 
}

\markboth{Journal of \LaTeX\ Class Files,~Vol.~14, No.~8, August~2021}%
{Anon \MakeLowercase{\textit{et al.}}: Reinforcement Learning Formulation for Real-World Environments via Off-Policy Evaluation}


\maketitle

\begin{abstract}
Reinforcement learning (RL) has the potential to transform real-world decision-making systems by enabling autonomous agents to learn from experience. However, deploying RL in real-world settings, especially in the context of human-robot interaction, requires defining appropriate state representations and reward functions, which are critical for learning efficiency and policy performance. Traditional RL approaches often rely on domain expertise and trial-and-error, necessitating extensive human involvement as well as direct interaction with the environment, which can be costly and impractical, especially in complex and safety-critical applications. This work proposes a novel RL framework that leverages off-policy evaluation (OPE) for state space and reward function selection, using only logged interaction data. This approach eliminates the need for real-time access to the environment or human-in-the-loop feedback, greatly reducing the dependency on costly real-time interactions. 
The proposed approach systematically evaluates multiple candidate state representations and reward functions by training offline RL agents and applying OPE to estimate policy performance. The optimal state space and reward function are selected based on their ability to produce high-performing policies under OPE metrics. Our method is validated on two environments: the Lunar Lander environment by OpenAI Gym, which provides a controlled setting for assessing state space and reward function selection, and a NASA-MATB-II human subjects study environment, which evaluates the approach's real-world applicability to human-robot teaming scenarios. This work enhances the feasibility and scalability of offline RL for real-world environments by automating critical RL design decisions through a data-driven OPE-based evaluation, enabling more reliable, effective, and sustainable RL formulation for complex human-robot interaction settings.
\end{abstract}

\section{Introduction}
\label{sec:intro}

Reinforcement learning (RL) has emerged as a powerful framework for optimizing sequential decision-making in dynamic environments. Notably, it has resulted in success in domains such as robotics, healthcare, and autonomous systems \cite{mnih2015human, levine2020offline, yu2021combo, liu2020reinforcement, zhu2022hierarchical} by enabling agents to learn optimal policies through interaction. However, real-world RL deployment remains challenging due to the complexities of defining effective state representations and reward functions; these significantly impact learning efficiency and policy performance \cite{dulac2020empirical, lesort2018state}. Traditional RL approaches require extensive online interactions, which can be costly, time-consuming, and unsafe \cite{prudencio2023survey, garcia2015comprehensive}. In human-robot interaction (HRI) scenarios, policies must align with human intentions while ensuring safety and adaptability \cite{shafti2020real}. Although offline RL mitigates these issues by training policies using pre-collected data \cite{levine2020offline, fu2020d4rl},  selecting appropriate state representations and reward functions without online experimentation remains a challenge.

Off-policy evaluation (OPE) provides a viable solution by enabling policy assessment using logged interaction data \cite{saito2021evaluating, uehara2022review}. OPE methods seek to estimate the expected performance of a policy without deploying it in the real-world environment, making them particularly useful for offline RL applications. While OPE has historically been employed to compare policies, its potential for guiding the selection of state spaces and reward functions remains underexplored.

This work proposes a novel RL formulation that leverages OPE for state space and reward function selection, enabling more effective RL deployment for real-world settings. By systematically evaluating multiple candidate state representations and reward functions using offline RL agents and OPE metrics, this approach identifies optimal configurations that yield high-performing policies. Crucially, our scheme eliminates the need for online evaluations, significantly reducing the cost and complexity of RL design, particularly for human-in-the-loop systems. Our method enhances RL feasibility by automating critical design choices through a structured, data-driven process, minimizing trial-and-error dependence. This is particularly valuable for safety-critical domains, e.g., healthcare, autonomous driving, and aerospace, given that the framework reduces the risk of deploying suboptimal policies, streamlining RL design and improving robustness.

The method that we propose is empirically validated in two RL environments: 
\textbf{1)} the OpenAI Gym Lunar Lander problem, a controlled testbed for assessing state and reward selection strategies \cite{brockman2016openai}, and 
\textbf{2)} a NASA-MATB-II human subjects study task set, which allows us to evaluate applicability to human-robot teaming \cite{gao2024off}. By leveraging OPE beyond traditional policy evaluation, this framework advances RL reliability, efficiency, and scalability for real-world problem settings. In essence, the \textbf{key contributions} of this research are:
\begin{itemize}[noitemsep,nolistsep]
    \item We introduced an OPE framework for selecting optimal state spaces and reward functions from logged data, bypassing real-time interactions; 
    \item We develop a method that optimizes state spaces and reward functions, ultimately improving performance in  controlled (Lunar Lander) and real-world (NASA-MATB-II) human-robot interaction scenarios.
\end{itemize}

The rest of this paper is organized in the following manner. 
Section \ref{sec:relatedwork} presents the background work and Section \ref{sec:method} outlines the proposed off-policy evaluation (OPE) framework for state and reward function selection. Next, Section \ref{sec:results} details the experimental setup, including the environments used (Lunar Lander and NASA-MATB-II) and the evaluation methodology as well as  presents the results of the experiments. Finally, Section \ref{sec:conclusion} concludes the article and discusses potential future directions for research.

\section{Background} 
\label{sec:relatedwork}

\subsection{Reinforcement Learning for Human-Robot Teaming}
\label{sec:rl_humanrobot_teaming}

RL has been shown to enable robots to adapt dynamically in human-robot collaboration by learning from human and environmental feedback; this adaptability is crucial in evolving tasks where human behavior is unpredictable \cite{akalin2021reinforcement, zhang2022reinforcement, thumm2024human}. However, real-world RL deployment faces challenges due to the high cost, time, and safety risks associated with extensive interactions required for optimal policy learning.

Offline reinforcement learning (ORL) addresses these issues by enabling policy learning from pre-collected data, reducing the need for real-time interactions and improving generalization across human behaviors \cite{levine2020offline, prudencio2023survey, wang2025provable}. However, ensuring offline-trained policies are safe and effective remains a challenge, making off-policy evaluation (OPE) essential. OPE estimates policy performance using logged data without additional interactions, which is critically useful in safety-sensitive applications \cite{prudencio2023survey, akalin2021reinforcement, saito2021evaluating, voloshin2019empirical}. Despite ORL and OPE advancements, evaluating offline-trained policies in dynamic environments remains difficult. OPE methods like importance sampling, fitted-Q evaluation, and doubly robust estimators address distributional shifts and high variance, making them essential tools for reliable policy evaluation \cite{kostrikov2020statistical, le2019batch, dudik2011doubly, gao2024trajectory}.

\subsection{Off-Policy Evaluation}
OPE methods are crucial for assessing the performance of a policy $\pi$ using data collected from a different behavior policy $\mu$, 
 making them essential in human-robot collaboration \cite{saito2021evaluating, gao2024off, voloshin2019empirical}. Again, this class of methods permits the evaluation of policies without additional real-world interaction, which is vital for use in safety-critical or costly real-time policy testing domains including healthcare, autonomous driving, and aerospace  A variety of OPE methods have been developed, each addressing different challenges. 
 Below, we outline some of the most widely used approaches.\cite{prudencio2023survey, akalin2021reinforcement}.



\begin{figure*}[!t]
\centering
\includegraphics[width=0.6\linewidth]{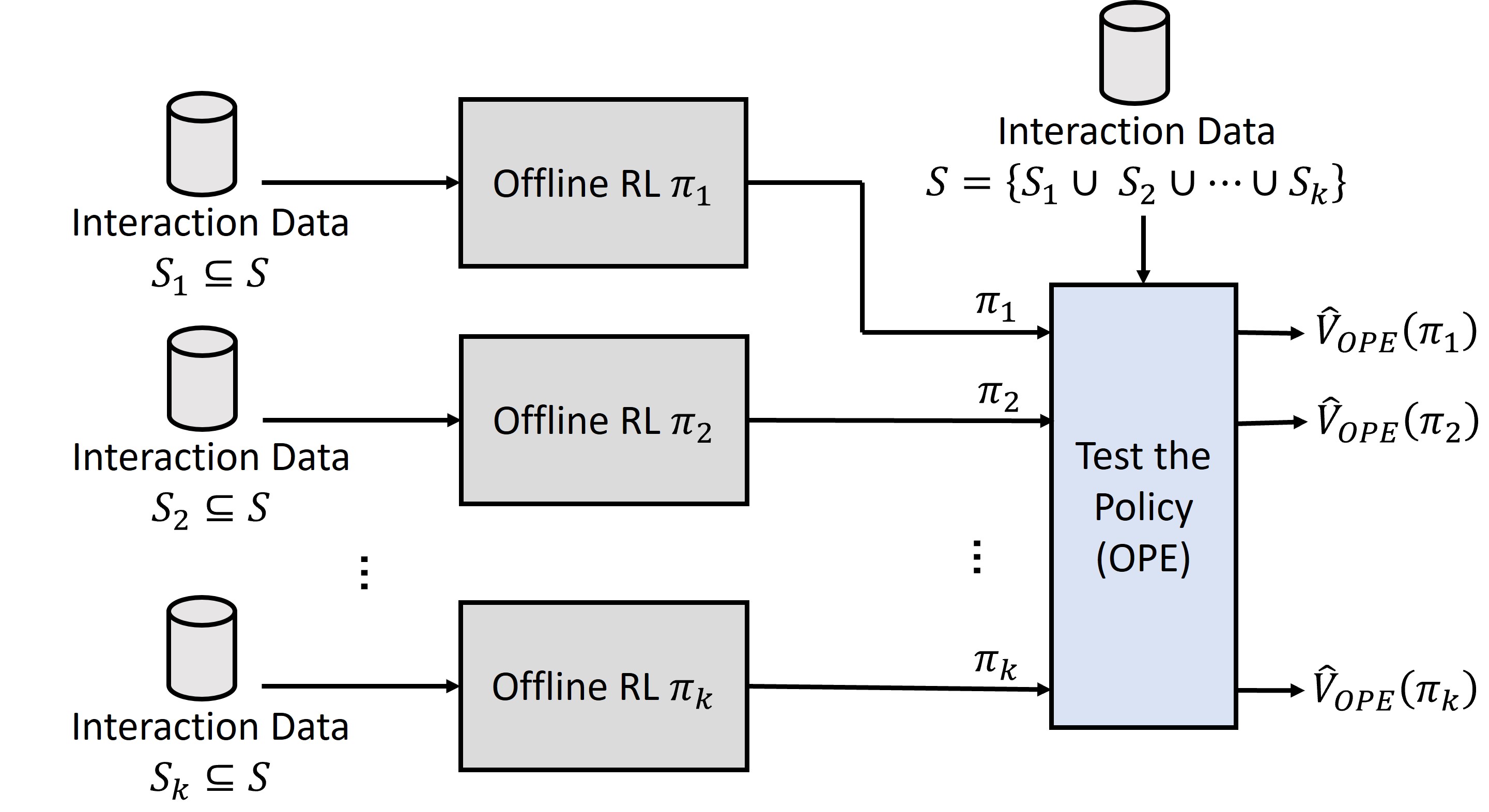}
\caption{The proposed scheme for state space selection using off-policy evaluation (OPE).}
\label{fig:OPE_state}
\end{figure*}

\textbf{Importance Sampling (IS)} is a model-free scheme for adjusting to the distribution shift between behavior and target policies by reweighting the collected data using importance weight values $w(x) = \frac{\pi(a_i | s_i)}{\mu(a_i | s_i)}$; this leads to an unbiased estimate of the target policy’s value. While effective, IS can suffer from high variance when policies are significantly different. Variants such as weighted importance sampling (WIS) and doubly robust estimators work to mitigate this issue by improving overall stability \cite{precup2000eligibility, dudik2011doubly, hanna2021importance}. On the other hand, the \textbf{direct method (DM)} estimates the reward function directly from data without requiring action probabilities. DM has lower variance than IS but can introduce bias if the reward model is inaccurate, especially in complex human-robot interactions \cite{abbeel2004apprenticeship, palan2019learning}.

In contrast to IS and DM, \textbf{fitted-Q evaluation (FQE)} learns a Q-function from offline data in order to estimate the target policy’s value. FQE is particularly useful in high-dimensional state spaces, such as robotics, where simpler methods often struggle \cite{le2019batch}. FQE is highly applicable to human-robot interaction tasks involving complex state spaces \cite{bobu2021feature, akalin2021reinforcement, zhang2022off}. Note that combining IS and DM results in the \textbf{doubly robust (DR)} estimator, reducing IS variance while  mitigating DM bias; DR remains consistent as long as either the reward model or importance weights are accurate \cite{dudik2011doubly, thomas2016data, funk2011doubly, shardell2024doubly}. Furthermore, DR is particularly effective in safety-critical domains such as healthcare and autonomous driving, ensuring robust policy evaluation before deployment \cite{saito2021evaluating}.

Recent work has explored improving OPE through trajectory augmentation, addressing challenges related to scarce and under-representative offline training trajectories. Gao et al. \cite{gao2024trajectory} proposed an offline trajectory augmentation approach that extracts valuable sub-trajectories and diversifies behavior coverage within those sub-trajectories. This method enhances OPE performance across various domains, including robotic control, healthcare, and e-learning, where data limitations typically pose the greatest challenge. OPE methods ultimately enable safe and efficient policy evaluation, providing robust RL policy assessments without requiring real-world testing. 

\section{Method}
\label{sec:method}

\subsection{State Space Selection via Off-Policy Evaluation}
\label{sec:method_state}

Selecting an appropriate state representation is crucial in real-world applications with costly data collection. This work introduces a framework leveraging OPE to systematically identify an optimal state space representation by evaluating candidate representations using logged interaction data (see Figure \ref{fig:OPE_state}). This extends OPE, beyond policy evaluation, to assess and rank multiple state space definitions based on their ability to induce high-performing policies \cite{shen2021state, hao2024off}.

\noindent
\textbf{Candidate State Space Definitions:} Given logged interaction data $\mathcal{D} = {(s_t, a_t, r_t, s_{t+1}, d_t)}_{t=1}^{N}$, we define candidate state representations $\mathcal{S}_1, \mathcal{S}_2, ..., \mathcal{S}_K$. Each representation $\mathcal{S}_k$ is derived from domain knowledge, feature selection techniques, or automated feature extraction \cite{pmlr-v235-wang24bh}. 

\noindent 
\textbf{Offline RL Policy Training:} An offline RL policy $\pi_k$ is trained using an offline RL algorithm for each candidate state space $\mathcal{S}_k$. The training objective is to optimize the expected return as follows: \\
$J(\pi_k) = \mathbb{E}{\tau \sim \pi_k} \left[ \sum_{t=0}^{T} \gamma^t r_t \right], 
$
where $\tau$ denotes a trajectory sampled under $\pi_k$ and $\gamma \in (0,1]$ is the discount factor.

\noindent 
\textbf{Off-Policy Evaluation:} The effectiveness of each state space representation is assessed using an OPE estimator $\hat{V}_{OPE}(\pi_k)$ trained on the complete dataset $\mathcal{D}$. The state space is defined as the union of all candidate representations: $\mathcal{S} = \{\mathcal{S}_1\cup \mathcal{S}_2\cup ...\cup \mathcal{S}_K\}$. The OPE method estimates the expected return of a given policy, without direct online interactions with the environment, in the following manner: 
${\hat{V}_{OPE}(\pi) = \mathbb{E}_{s \sim d_\beta, a \sim \pi} \left[ Q^{\pi}(s, a) \right], }$ where $d_\beta$ represents the state distribution under the behavior policy $\beta$. Common OPE techniques, such as importance sampling (IS), fitted Q-evaluation (FQE), or model-based approaches, may be employed to compute $\hat{V}_{OPE}(\pi)$ \cite{le2019batch, dudik2011doubly, irpan2019off}.

\noindent
\textbf{Ranking and Selection:} The selection of the optimal state space representation is essentially performed by ranking the candidate state spaces based on their corresponding OPE-estimated policy performance via: ${\mathcal{S}^* = \arg\max_{\mathcal{S}_k} \hat{V}_{OPE}(\pi_k), }$ where $\mathcal{S}^*$ leads to the most performant offline RL policy under OPE evaluation. This methodology facilitates principled, data-driven state space selection, further eliminating costly trial-and-error interactions in real-world RL deployment.

\subsection{Reward Function Selection via Off-Policy Evaluation}
\label{sec:method_reward}

\begin{figure*}[!t]
\centering
\includegraphics[width=0.75\linewidth]{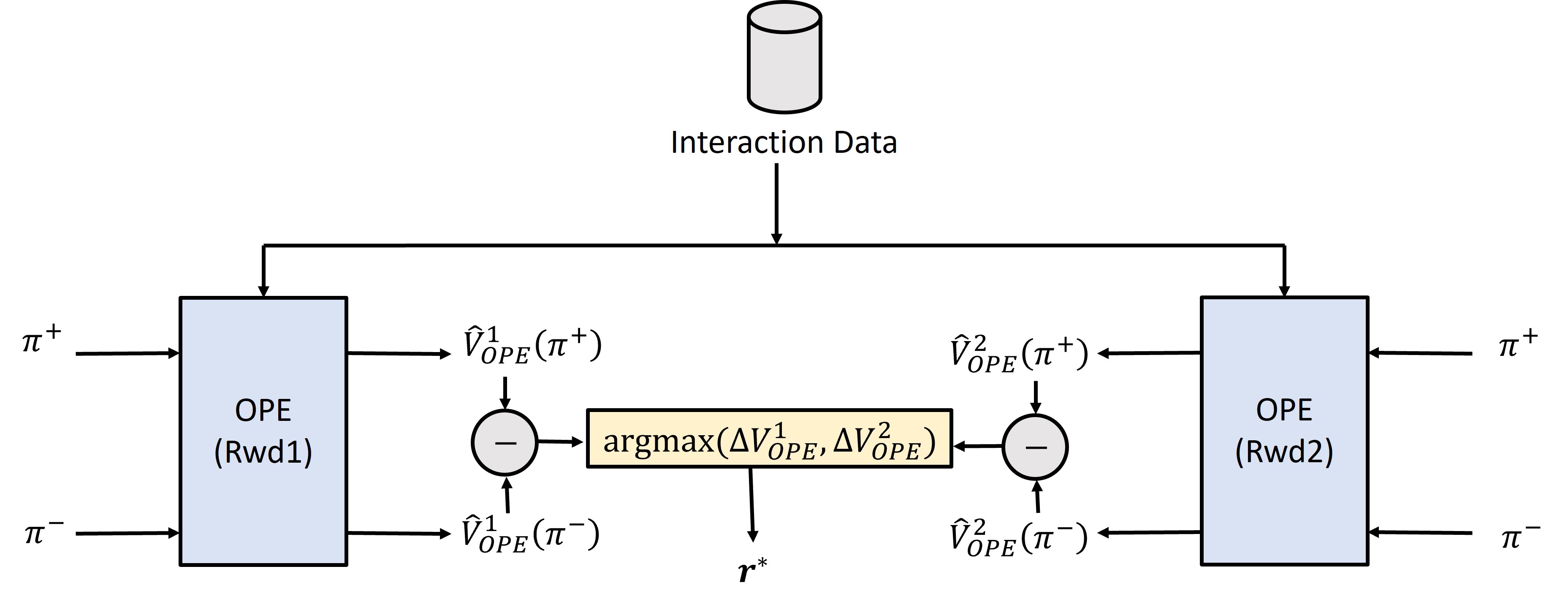}
\caption{Our process for selecting a reward function based on the separability of good and bad policies using OPE.}
\label{fig:OPE_rewards}
\end{figure*}

Defining an effective reward function is critical for RL in real-world applications. This section presents a structured approach for utilizing OPE to compare multiple reward functions based on their ability to distinguish high- and low-performing policies; see Figure \ref{fig:OPE_rewards}.

\noindent
\textbf{Candidate Reward Function Definitions:} Given logged interaction data $\mathcal{D} = {(s_t, a_t, r_t^j, s_{t+1}, d_t)}_{t=1}^{N}$ for each candidate reward function $r^j$, we define a set of reward functions ${r^1, r^2, ..., r^M}$. These are derived from domain knowledge and are structured to capture different characteristics of a task; outputs are normalized for comparability across policies.

\noindent 
\textbf{Training Off-Policy Evaluators:} For each reward function $r^j$, an OPE estimator $\hat{V}_{OPE}^{j}(\pi)$ computes, as mentioned before, the expected return under policy $\pi$ via: ${\hat{V}_{OPE}^{j}(\pi) = \mathbb{E}_{s \sim d\beta, a \sim \pi} \left[ Q^{j,\pi}(s, a) \right]}$, where $Q^{j,\pi}(s,a)$ is the action-value function of policy $\pi$ under reward function $r^j$.

\noindent 
\textbf{Evaluating Good and Bad Policies:} A predefined high-performing policy $\pi^+$ and a low-performing policy $\pi^-$ are evaluated under each reward function. Note that a well-defined reward function should induce a clear separation between their estimated returns: ${\Delta V_{OPE}^{j} = \hat{V}_{OPE}^{j}(\pi^+) - \hat{V}_{OPE}^{j}(\pi^-)}$, where $\Delta V_{OPE}^{j}$ quantifies the ability of $r^j$ to distinguish between good and bad policies.

\noindent 
\textbf{Quantifying Reward Function Effectiveness:} Three divergence measures are computed on standardized policy returns to assess how well each reward function separates policy performance. Specifically, the mean return distance ($\Delta V_{OPE}^{j}$), the Kullback-Leibler (KL) divergence, and the Jensen-Shannon (JS) divergence. Note that each return distribution was standardized for fair comparisons.

\textbf{Selecting the Optimal Reward Function:} The final reward function is selected as follow: ${r^* = \arg\max_{r^j} \Delta V_{OPE}^{j} }$, where $r^*$ maximizes policy performance separation, ensuring a strong learning signal for RL agents. Our approach enables systematic reward shaping using OPEs, reducing the need for costly online interactions. Thus, the method enhances RL reliability in real-world applications by ensuring that the chosen reward function effectively differentiates policy performance. 

\section{Offline Evaluation}
\label{sec:results}

The proposed method is validated in two environments: 
\textbf{1)} the OpenAI Gym Lunar Lander \cite{brockman2016openai}, 
and 
\textbf{2)} a NASA-MATB-II human subjects study \cite{gao2024off}. 

\subsection{Lunar Lander Environment}
\label{sec:results_lunar_lander}

\begin{figure}[htbp]
\centering
\includegraphics[width=\linewidth]{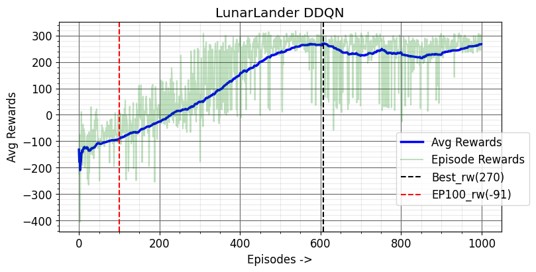}
\caption{Average rewards over $1000$ episodes for a DDQN agent trained on the Lunar Lander environment. The model, at episode $100$, is used as the \textit{Avg.} DDQN model, whereas the model at episode $607$ is used as the \textit{Best} DDQN model.}
\label{fig:lunarlander}
\end{figure}

The Lunar Lander environment serves as a testbed to validate our proposed methodology by evaluating its state space representation and reward function selection capability. Different double deep Q-network (DDQN) agents were trained in the Lunar Lander environment for over $1000$ episodes in order to collect interaction data and to test the general performance of OPEs on the Lunar Lander environment itself; see Figure \ref{fig:lunarlander}. The DDQN agent was trained with discount factor $\gamma=0.99$, learning rate of $0.00005$, batch size of $128$, and target network interpolation rate of $\tau=0.01$. All neural networks contained $2$ hidden layers with $256$ neurons each with ReLU activation functions and Gaussian initialization. 

\begin{figure}[htbp]
\centering
\includegraphics[width=0.7\linewidth]{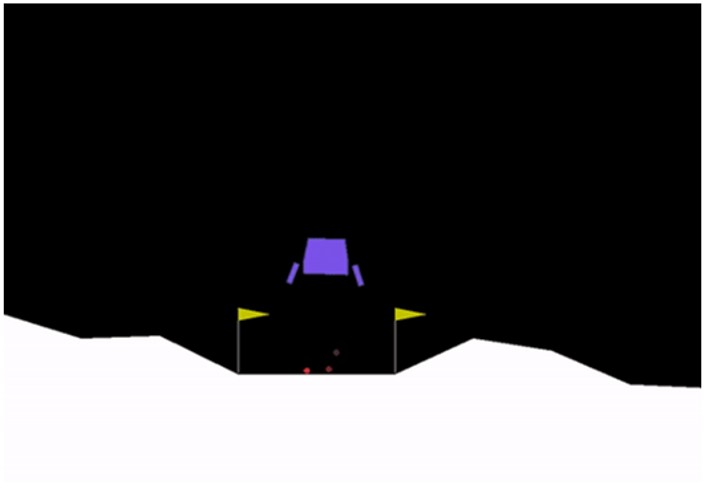}
\caption{Lunar Lander Environment.}
\label{fig:lunarlanderenv}
\end{figure}

\begin{table*}[b]
\centering
\caption{Off-Policy Evaluation performance on Lunar Lander}
\label{tab:ope_perf}
\begin{tabular}{|c|ccc|ccc|}
\hline
\multirow{2}{*}{\textbf{OPE}} & \multicolumn{3}{c|}{\textbf{Online Models (DDQN)}} & \multicolumn{3}{c|}{\textbf{Offline Models (CQL)}} \\ \cline{2-7} 
 & \multicolumn{1}{c|}{\textbf{Random}} & \multicolumn{1}{c|}{\textbf{Avg.}} & \textbf{Best} & \multicolumn{1}{c|}{\textbf{Worst}} & \multicolumn{1}{c|}{\textbf{Avg.}} & \textbf{Best} \\ \hline
Ground Truth & \multicolumn{1}{c|}{-153.64} & \multicolumn{1}{c|}{-113.20} & 139.05 & \multicolumn{1}{c|}{-211.01} & \multicolumn{1}{c|}{-9.48} & 119.40 \\ \hline
Importance Sampling & \multicolumn{1}{c|}{53.39} & \multicolumn{1}{c|}{59.43} & 62.96 & \multicolumn{1}{c|}{59.48} & \multicolumn{1}{c|}{65.07} & 67.64 \\ \hline
Direct Method FQE & \multicolumn{1}{c|}{9225.93} & \multicolumn{1}{c|}{9236.98} & 9244.85 & \multicolumn{1}{c|}{9229.34} & \multicolumn{1}{c|}{9241.31} & 9247.28 \\ \hline
\end{tabular}
\end{table*}

DDQN models with average and best performance were selected to assess the effectiveness of the OPE methods. This interaction data was then used to train offline RL agents using constrained Q-learning (CQL). These offline agents were trained with varying amounts of data to produce worst, average, and best-performing policies. Importance sampling (IS) and the direct method (DM) OPE methods were applied to evaluate the models. As shown in Table \ref{tab:ope_perf}, both IS and DM correctly ranked the policy returns, with the best-performing policy achieving the highest return and the worst-performing policy the lowest.

\subsubsection{State Space Selection}
An experiment was conducted to determine how effective the OPEs were at selecting state spaces offline. Three candidate state space representations were investigated: original, expanded, and reduced. The original state space ($S_{orig}$) consisted of $\{x, y, \dot{x}, \dot{y}, \theta, \dot{\theta}, L_{left}, L_{right}\}$. The expanded states space ($S_{more}$) added two zero-mean Gaussian noise variables to the original state space; in contrast, the reduced state space ($S_{less}$) omitted position variables ($x, y$).



CQL agents were trained on each candidate representation and evaluated using the hybrid direct method with fitted Q-evaluation (DM-FQE) OPE, trained on the union of all state spaces ($S_{orig} \cup S_{more} \cup S_{less}$). Results, shown in Table \ref{tab:lunarlander_state}, indicate that $S_{orig}$ led to the highest-performing policy under OPE evaluation, reinforcing that unnecessary feature expansion can degrade performance while excessive compression omits critical information. This demonstrates that unnecessary feature expansion can degrade performance while a reduction in representative features can degrade the estimation. From this, we may observe that OPE methods uncover the feature relevance of the agent's state space. 

\begin{table}[htbp]
\centering
\caption{State space selection via the direct method with fitted Q-evaluation (DM-FQE) OPE results for the Lunar Lander environment.}
\label{tab:lunarlander_state}
\begin{tabular}{|c|c|c|}
\hline
\textbf{\begin{tabular}[c]{@{}c@{}}Candidate \\ State Space\end{tabular}} & \textbf{\begin{tabular}[c]{@{}c@{}}Online Eval. Rewards\\ ($\sum_{t=0}^{T} r_t$)\end{tabular}} & \textbf{\begin{tabular}[c]{@{}c@{}}OPE Estimate Return\\ ($\hat{V}_{OPE}(\pi_k)$)\end{tabular}} \\ \hline
$S_{orig}$ & \textbf{230.86 (60.73)} & \textbf{8954.86} \\ \hline
$S_{more}$ & 93.96 (145.18) & 8950.66 \\ \hline
$S_{less}$ & -15.47 (22.71) & 8947.54 \\ \hline
\end{tabular}
\end{table}

\subsubsection{Reward Function Selection}
The capabilities of OPE for reward selection were also validated using the Lunar Lander environment. The baseline reward function $f(s,a)$ incorporates short-horizon state-based, action-based, and terminal-state rewards. Three alternative reward functions were synthesized: 
\textbf{1)} $f_{r1}(s,a)$: terminal-state rewards only; 
\textbf{2)} $f_{r2}(s,a)$: action-based and terminal-state rewards; and,  
\textbf{3)} $f_{r3}(s,a)$: state-based and terminal-state rewards.


CQL agents, trained using each reward function, were evaluated via the DM-FQE OPE. The OPE estimated values, for each reward function, were analyzed by measuring their ability to distinguish between good and poor policy performance. Table \ref{tab:lunarlanderrwdsel} summarizes these results, showing that $f(s,a)$ exhibits the highest Jensen-Shannon (JS) divergence, signifying superior policy separability. This further highlights the importance of incorporating multiple reward components for effective learning signals. These results validate the proposed OPE-based selection framework, demonstrating its utility in identifying optimal state representations and reward functions without requiring costly trial-and-error interactions. 


\begin{table}[htbp]
\centering
\caption{Reward function selection via the direct method with fitted Q-evaluation (DM-FQE) off-policy evaluation for the Lunar Lander environment. CQL agents with state space $S_{orig}$ were used as the best and worst policies.}
\label{tab:lunarlanderrwdsel}
\resizebox{\linewidth}{!}{%
\begin{tabular}{|c|c|c|c|c|}
\hline
\textbf{Reward Function} & \textbf{$f(s,a)$} & \textbf{$f_{r1}(s,a)$} & \textbf{$f_{r2}(s,a)$} & \textbf{$f_{r3}(s,a)$} \\ \hline
\textbf{Mean Worst Policy} & 7.67e-16 & 0.0 & 1.13e-16 & -1.13e-16 \\ \hline
\textbf{Mean Best Policy} & 6.53e-16 & -1.42e-16 & 8.52e-17 & 3.69e-16 \\ \hline
\textbf{Mean Difference} & -1.13e-16 & -1.42e-16 & -2.84e-17 & 4.83e-16 \\ \hline
\textbf{JS Divergence} & \textbf{3.56e-5} & 2.00e-5 & 9.11e-6 & 2.75e-5 \\ \hline
\textbf{KL Divergence} & 1.65e-4 & -1.38e-4 & 4.18e-4 & 1.63e-4 \\ \hline
\end{tabular}%
}
\end{table}

\subsection{NASA MATB-II Human Subject's Study}
\label{sec:results_matbii}

A human-subjects study was conducted using the NASA Multi-Attribute Task Battery-II (MATB-II) \cite{Comstock1992} in order to assess the real-world applicability of our proposed RL framework in the context of human-robot teaming. MATB-II is a widely used simulation platform in human factors and cognitive workload research, designed to emulate dynamic, multitasking operational environments. It usefully enables precise manipulation of workload conditions while capturing behavioral and physiological responses during task performance \cite{cegarra2020openmatb}. This study evaluates how offline RL agents, trained and selected using the proposed OPE-based framework, compare against rule-based automation in adapting to human cognitive states and effectively supporting task performance. The experiment serves as a testbed for validating the effectiveness of data-driven RL design in human-in-the-loop settings by leveraging MATB-II’s complex task structure and rich multimodal data \cite{vogl2024united, cegarra2020openmatb, gugerell2024studying}. 

MATB-II evaluates how individuals manage multiple tasks simultaneously, such as monitoring systems, responding to alerts, and performing manual controls. MATB-II comprises four concurrent tasks:
\begin{itemize}[noitemsep,nolistsep]
\item \textbf{Tracking Task:} Use a joystick to maintain a crosshair on a moving target. Performance is measured as the root mean squared error (RMSE) and the mean absolute error (MAE) between the target and crosshair positions, indicating the accuracy of target tracking. 
\item \textbf{System Monitoring Task:} Monitor four gauges and two alarm lights, resetting them when they go out of range. Performance is assessed based on the hit rate (percentage of system failures correctly identified), false alarm rate (percentage of incorrect responses when no failure occurred), and response time (time taken to acknowledge a system failure).
\item \textbf{Resource Management Task:} Regulate fuel levels in tanks A and B by adjusting pump settings to maintain fuel levels within the optimal range ($2000$–$3000$ units). Performance is measured by the mean absolute deviation (MAD) of fuel levels from the target, the standard deviation (SD) of fuel level maintenance, and the percentage of time both tanks remain within the designated range.
\item \textbf{Communications Task:} Listen to and respond to audio commands, such as adjusting frequencies based on verbal instructions. Performance is measured in terms of the correct response rate (percentage of correct responses to auditory instructions), reaction time (time taken to respond to communications), and the error rate (percentage of incorrect or missed responses).
\end{itemize}

\begin{figure}[htbp]
\centering
\includegraphics[width=\linewidth]{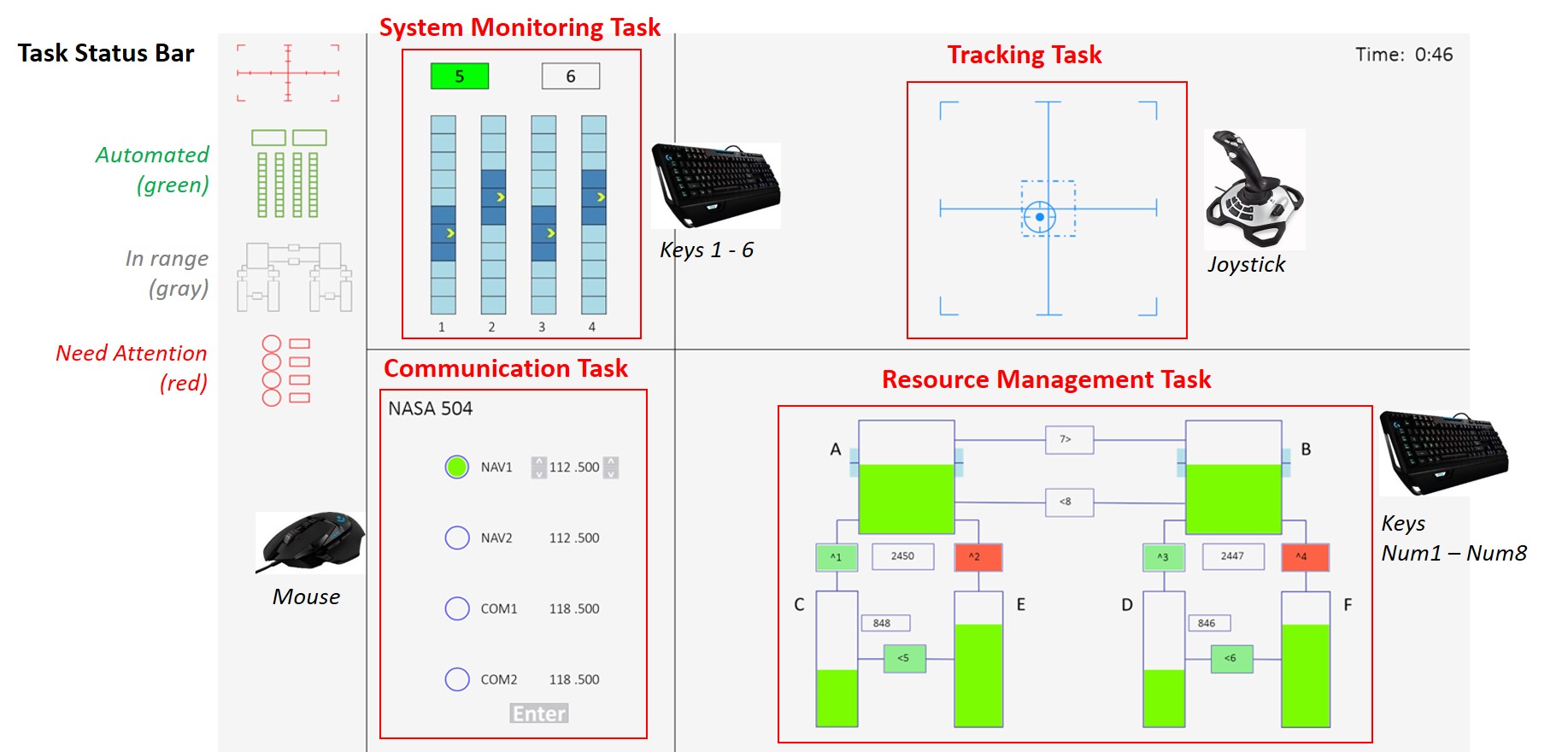}
\caption{The NASA Multi-Attribute Task Battery-II (MATB-II) Environment.}
\label{fig:nasa_matb}
\end{figure}

Figure \ref{fig:nasa_matb} illustrates the MATB-II interface. To facilitate automation, modifications were introduced to allow each task to be independently automated, ensuring near-perfect task execution. The left-side status bar in Figure \ref{fig:nasa_matb} visually communicates task states: red (requires attention), gray (parameters within range), and green (task automated).

Data was collected from a prior human-subject study \cite{singh2022human}\cite{singhheard2022humanHRI}
involving nine participants ($5$ males and $4$ females, with average age of $26.3$) under varying workload conditions (`Underload', `Normal Load', `Overload'). Participants interacted with the MATB-II environment, with automation managed by adaptive agents: rule-based (RB) and RL-based. Data collected included physiological measures, e.g., heart rate, respiration, task interaction metrics, workload estimates \cite{Heard2019thri, fortune2020real, heard2017human}, and subjective assessments (NASA-TLX \cite{hart1988development}). This study builds upon this dataset by implementing an RL framework for automation, using off-policy evaluation (OPE) and offline RL training. Several state-space configurations and reward functions are analyzed to identify the most effective set-up for learning policies. The RL agent trained on the best state-space and reward function was evaluated in an online experiment. 

The dataset used consists of three types of information: 
\textit{(i)} task-related information, 
\textit{(ii)} physiological data, and 
\textit{(iii)} estimated workload data. These three data types were combined in different ways to form various candidate state-space representations, as shown in Table \ref{tab:nasamatbstatespace}. The reward function candidates are shown in Table \ref{tab:nasamatbrewardfunc}. The analysis focused on two objectives: 
\textit{(1)} evaluating the effectiveness of state-space representations for RL-based automation, and 
\textit{(2)} selecting the optimal reward function for training RL agents.

\begin{table}[htbp]
\centering
\caption{State space representation candidates for the NASA-MATB-II task environment.}
\label{tab:nasamatbstatespace}
\begin{tabular}{|l|c|c|c|}
\hline
\multicolumn{1}{|c|}{\textbf{\begin{tabular}[c]{@{}c@{}}State Space\\ Definition\end{tabular}}} & \textbf{\begin{tabular}[c]{@{}c@{}}Task\\ Information\end{tabular}} & \textbf{\begin{tabular}[c]{@{}c@{}}Physiological\\ Data\end{tabular}} & \textbf{\begin{tabular}[c]{@{}c@{}}Workload\\ Data\end{tabular}} \\ \hline
$S_{task+physio+wl}$ & \ding{51} & \ding{51} & \ding{51} \\ \hline
$S_{task+physio}$ & \ding{51} & \ding{51} &  \\ \hline
$S_{task+wl}$ & \ding{51} &  & \ding{51} \\ \hline
$S_{task}$ & \ding{51} &  &  \\ \hline
$S_{physio}$ &  & \ding{51} &  \\ \hline
$S_{wl}$ &  &  & \ding{51} \\ \hline
\end{tabular}
\end{table}

\begin{table*}[htbp]
\centering
\caption{Reward function candidates for the NASA-MATB-II task environment.}
\label{tab:nasamatbrewardfunc}
\begin{tabular}{|l|l|l|}
\hline
\multicolumn{1}{|c|}{\textbf{\begin{tabular}[c]{@{}c@{}}Reward Function\\ Definition\end{tabular}}} & \multicolumn{1}{c|}{\textbf{Description}} & \multicolumn{1}{c|}{\textbf{Reward Function Components}} \\ \hline
\multirow{6}{*}{$r_1(s,a)$} & \multirow{6}{*}{\begin{tabular}[c]{@{}l@{}}Penalize bad\\ behavior\end{tabular}} & \begin{tabular}[c]{@{}l@{}}Sysmon: - a*(\# gauges out of range)\\ \hspace{1.4cm} - b*(\# lights out of range)\end{tabular} \\ \cline{3-3} 
 &  & Trck: - c*(tracking error) \\ \cline{3-3} 
 &  & \begin{tabular}[c]{@{}l@{}}Resman: (0 if TA in range else -d*(Out of range error))\\ \hspace{1.4cm}+ (0 if TB in range else -d*(Out of range error))\end{tabular} \\ \cline{3-3} 
 &  & \begin{tabular}[c]{@{}l@{}}Comm: - e*max(comm\_event\_timeout-8, 0)\\ \hspace{1.4cm}*(comm\_radio\_ok + comm\_frew\_ok)/2\end{tabular} \\ \cline{3-3} 
 &  & Human Idle: - f if human idle else 0 \\ \cline{3-3} 
 &  & Automation: - g if automation \\ \hline
\multirow{6}{*}{$r_2(s,a)$} & \multirow{6}{*}{\begin{tabular}[c]{@{}l@{}}Reward good\\ behavior\end{tabular}} & Sysmon: + h (for fixing a gauge or light) \\ \cline{3-3} 
 &  & Trck: + i*max((max\_err/2 - tracking error), 0) \\ \cline{3-3} 
 &  & \begin{tabular}[c]{@{}l@{}}Resman: + (j if TA in range else 0)\\ \hspace{1.4cm}+ (j if TB in range else 0)\end{tabular} \\ \cline{3-3} 
 &  & Comm: + k * (comm\_radio\_ok + comm\_frew\_ok)/2 \\ \cline{3-3} 
 &  & Human Idle: +l if not human idle else 0 \\ \cline{3-3} 
 &  & Automation: + m if no automation \\ \hline
$r_3(s,a)$ & \begin{tabular}[c]{@{}l@{}}Weighted sum of\\ $r_1(s,a)$ and $r_2(s,a)$\end{tabular} & p*$r_1(s,a)$+ q*$r_2(s,a)$ \\ \hline
\end{tabular}%
\end{table*}

\subsubsection{State-Space Representation Selection}
\label{sec:state_space_rep_selection}

OPE results using the direct method with fitted Q-evaluation (DM-FQE) for various state-space representations are presented in Table \ref{tab:nasamatbstatesel}. Observe that the multimodal state representation $S_{task+physio+wl}$ achieved the highest OPE return ($58520030.82$), indicating that combining task, physiological, and workload data provides the most informative state for RL. The second-best configuration, $S_{task+physio}$ ($58150584.05$), indicates that workload data may add only marginal benefit when task and physiological data are already included. Note that $S_{wl}$ alone yielded the lowest return ($57122466.86$), confirming that workload data alone is insufficient to facilitate effective learning.

\begin{table}[htbp]
\centering
\caption{State Space Selection via Direct Method (DM-FQE) Off-Policy Evaluation for the NASA MATB-II environment. Note that reward function $r_3(s,a)$ is used for this analysis.}
\label{tab:nasamatbstatesel}
\begin{tabular}{|l|l|}
\hline
\textbf{Candidate State Space} & \textbf{\begin{tabular}[c]{@{}l@{}}OPE Estimate Return\\ ($\hat{V}_{OPE}(\pi_k)$)\end{tabular}} \\ \hline
$S_{task+physio+wl}$ & \textbf{58520030.82} \\ \hline
$S_{task+physio}$ & 58150584.05 \\ \hline
$S_{task}$ & 57959191.05 \\ \hline
$S_{task+wl}$ & 57917962.11 \\ \hline
$S_{physio}$ & 57225790.66 \\ \hline
Random & 57144435.09 \\ \hline
$S_{wl}$ & 57122466.86 \\ \hline
\end{tabular}
\end{table}

The state utilization (SU) analysis \cite{singh2024measuring} we further employed in our analysis shows that task-related data had the highest utilization across all configurations ($70.9$\%), with physiological and workload data contributing $19.1$\% and $10.1$\%, respectively. This supports the conclusion that task-related data is crucial for RL performance, whereas physiological and workload data are secondary.

\begin{table}[htbp]
\centering
\caption{State utilization (SU) for CQL agents trained on multimodal state space representations. Note that the reward function $r_3(s,a)$ is used for this analysis. ``-'' indicates that the modality was not present.}
\label{tab:su}
\begin{tabular}{|c|c|c|c|}
\hline
\textbf{ORL Agent} & \textbf{SU Task Info} & \textbf{SU Workload} & \textbf{SU Physio} \\ \hline
task+physio+wl & \textbf{0.709} & 0.101 & 0.191 \\ \hline
task+physio & \textbf{0.751} & - & 0.249 \\ \hline
task+wl & \textbf{0.863} & 0.137 & - \\ \hline
\end{tabular}
\end{table}

Our findings show that multimodal state representations can enhance RL-based automation. However, due to the minimal additional benefit and computational overhead of workload data, $S_{task+physio}$ was chosen as the preferred state representation, with task-related data being the most important, followed by physiological information, and workload (which contributed the least to policy learning).

\subsubsection{Reward Function Selection}
The performance of different reward functions ($r_1(s,a)$, $r_2(s,a)$, and $r_3(s,a)$) was assessed using DM-FQE OPE; results summarized in Table \ref{tab:nasamatbrwdsel}. The evaluation compared the mean returns of a random policy with the best RL policy for each reward function, analyzing their differences and divergence metrics, i.e., Jensen-Shannon (JS) and Kullback-Leibler (KL) divergences.

All reward functions led to significant performance differences between the random and best policies, indicating effective RL training. Among them, $r_2(s,a)$ achieved the highest mean difference ($2.17$), showing strong policy separation. $r_3(s,a)$ also performed well, with a mean difference of $2.14$, the highest KL divergence ($9.52$), and the highest JS divergence ($0.65$), indicating the most distinguishable policy distribution. Our methodology's resulting balanced nature, combining penalties for undesirable actions and rewards to engender desirable behavior, makes it the most effective for training RL agents.


\begin{table}[htbp]
\centering
\caption{Reward function selection via the direct method with fitted Q-evaluation (DM-FQE) OPE for the NASA-MATB-II environment. A CQL agent with state space $S_{task+physio+wl}$ was used as the best policy.}
\label{tab:nasamatbrwdsel}
\begin{tabular}{|c|c|c|c|}
\hline
\textbf{Reward Function} & \textbf{$r_1(s,a)$} & \textbf{$r_2(s,a)$} & \textbf{$r_3(s,a)$} \\ \hline
\textbf{Mean Worst Policy} & -1.01 (0.31) & -1.16 (0.35) & -1.10 (0.29) \\ \hline
\textbf{Mean Best Policy} & 1.08 (0.49) & 1.01 (0.48) & 1.04 (0.42) \\ \hline
\textbf{Mean Difference} & 2.10 & \textbf{2.17} & 2.14 \\ \hline
\textbf{JS Divergence} & 0.64 & 0.62 & \textbf{0.65} \\ \hline
\textbf{KL Divergence} & 5.51 & 8.39 & \textbf{9.52} \\ \hline
\end{tabular}
\end{table}

Our findings confirm $r_3(s,a)$ as the most effective reward function for RL automation in NASA MATB-II. Our methodological framework enables agents to learn better policies more effectively while maintaining robust performance under varying workload conditions by combining both positive and negative reinforcement.

\section{Online Evaluation}
\label{sec:online_eval_results}

The analysis of the NASA MATB-II environment suggests that $S_{task+physio}$ as the state space and reward function $r_3(s,a)$ results in a high-performing ORL agent. This result is further validated through an IRB-approved human subjects study evaluating the selected state-space, reward function, and ORL agent from Section \ref{sec:method}. Nineteen participants ($10$ males, $9$ females; mean age: $23.1$) completed multiple sessions using the NASA MATB-II simulation platform. On a $1$–$5$ scale, the participants self-reported a mean video game skill level of $2.42$  and standard deviation (SD) of $1.07$, mean fatigue level of $1.84$ (SD $= 0.76$), and mean stress level of $1.73$ (SD $= 0.87$). Participants were exposed to three workload conditions -- `Underload' (UL), `Normal Load' (NL), and `Overload' (OL) -- to assess the RL agent's adaptability under varying cognitive demands. Each participant first completed a $15$-minute training session on the NASA MATB-II environment. This was followed by two $52.5$-minute sessions, during which the particpants interacted with either a rule-based (RB) agent or an RL or RL fine-tuning (RLFT) agent (the selection of RL or RLFT was random). The order of these sessions was randomized to mitigate ordering effects. Within each $52.5$-minute session, participants cycled through seven workload conditions (OL-UL-OL-NL-UL-NL-OL), with each condition lasting $7.5$ minutes.

Three types of agents were employed to evaluate the impact of different automation strategies:
\begin{itemize}[noitemsep,nolistsep]
    \item \textbf{Rule-Based (RB) Agent:} This agent automates tasks based on a set of predefined rules. It further monitors the human’s estimated workload and activates automatically when the human is overloaded. During overload conditions, the agent automates the task that is farthest from the one the human is currently engaged with.
    \item \textbf{Reinforcement Learning (RL) Agent:} Trained using an offline RL framework on the offline dataset, this agent learns optimal policies based on task-related, physiological, and workload data.
    \item \textbf{Reinforcement Learning Fine-Tuning (RLFT) Agent:} This agent builds on the \textit{RL agent} model and further fine-tunes or co-trains its policy using new data collected during the RLFT trial. By adapting its policy based on the participant’s real-time interactions, the RLFT agent enables more dynamic and responsive automation.
\end{itemize}

Physiological data, including heart rate, heart rate variability, respiration, posture, ambient noise levels, and speech pitch and intensity, were recorded to monitor cognitive states and to estimate objective workload during task performance. These measures are widely recognized for their effectiveness in estimating cognitive workload using machine learning and multimodal fusion techniques \cite{das2024cognitive, wei2025cognitive, diarra2025systematic}. Additionally, in-situ questions about workload, trust, and fluency were asked after each workload condition. Note that subjective workload, trust, and fluency were further assessed using the NASA-TLX questionnaire \cite{hart1988development}, the Trust in Automation questionnaire \cite{korber2019theoretical}, and the Fluency questionnaire \cite{hoffman2019evaluating}. Finally, task performance metrics were compared across the three agents to determine whether the RL agents could achieve better results in multi-task environment under different workload conditions compared to a rule-based approach.

\begin{table*}[hb]
\centering
\caption{Individual task performance in NASA MATB-II, averaged (standard deviation) over $19$ participants. Note: Lower is Better.}
\label{tab:nasamatb_online}
\begin{tabular}{|c|c|c|c|c|c|c|}
\hline
\textbf{Task} & \textbf{Performance Metric} & \textbf{Trial} & \textbf{Underload} & \textbf{Normal Load} & \textbf{Overload} & \textbf{Overall} \\ \hline
\multirow{3}{*}{Tracking} & \multirow{3}{*}{\begin{tabular}[c]{@{}c@{}}Average Root Mean \\ Squared Error (Pixels)\end{tabular}} & RB & \textbf{18.48 (3.79)} & 24.28 (4.37) & 31.86 (7.34) & 26.18 (4.65) \\
 &  & RL & 20.50 (3.12) & \textbf{23.97 (2.49)} & \textbf{26.36 (2.85)} & \textbf{24.10 (2.15)} \\
 &  & RLFT & 21.44 (7.21) & 25.71 (4.30) & \textbf{26.22 (1.67)} & \textbf{24.73 (3.14)} \\ \hline
\multirow{6}{*}{\begin{tabular}[c]{@{}c@{}}System \\ Monitoring\end{tabular}} & \multirow{3}{*}{\begin{tabular}[c]{@{}c@{}}Average Response \\ Time (Sec.)\end{tabular}} & RB & 8.62 (23.32) & 17.59 (12.13) & 11.89 (7.04) & 12.36 (6.70) \\
 &  & RL & \textbf{2.76 (1.20)} & \textbf{8.35 (10.41)} & \textbf{9.30 (4.23)} & \textbf{9.12 (4.10)} \\
 &  & RLFT & 17.18 (22.75) & 38.11 (43.40) & 14.83 (6.39) & 16.99 (7.62) \\ \cline{2-7} 
 & \multirow{3}{*}{Failure Rate (\%)} & RB & 5.26 (12.49) & 14.82 (9.08) & 20.15 (14.24) & 19.45 (13.26) \\
 &  & RL & \textbf{0.0 (0.0)} & \textbf{7.89 (10.02)} & \textbf{16.59 (10.85)} & \textbf{15.56 (10.04)} \\
 &  & RLFT & 10.00 (11.65) & 22.63 (17.90) & 27.24 (10.97) & 26.57 (10.40) \\ \hline
\multirow{3}{*}{\begin{tabular}[c]{@{}c@{}}Resource \\ Management\end{tabular}} & \multirow{3}{*}{Failure Rate (\%)} & RB & \textbf{6.28 (5.19)} & \textbf{1.58 (3.46)} & \textbf{6.88 (8.01)} & \textbf{5.41 (5.15)} \\
 &  & RL & 9.08 (11.94) & 6.80 (16.03) & 12.93 (18.36) & 10.23 (15.09) \\
 &  & RLFT & 7.32 (4.90) & 10.60 (18.16) & 13.66 (19.93) & 11.09 (14.02) \\ \hline
\multirow{6}{*}{Communication} & \multirow{3}{*}{\begin{tabular}[c]{@{}c@{}}Average Response \\ Time (Sec.)\end{tabular}} & RB & - & 12.29 (2.77) & 14.95 (3.08) & 14.57 (2.81) \\
 &  & RL & - & \textbf{10.96 (2.02)} & 15.45 (3.53) & 14.81 (3.13) \\
 &  & RLFT & - & \textbf{10.66 (2.57)} & \textbf{13.90 (3.47)} & \textbf{13.44 (3.20)} \\ \cline{2-7} 
 & \multirow{3}{*}{Failure Rate (\%)} & RB & - & 15.31 (12.88) & 25.92 (16.95) & 24.40 (15.02) \\
 &  & RL & - & 12.12 (14.37) & 27.78 (17.54) & 25.54 (15.99) \\
 &  & RLFT & - & \textbf{7.27 (9.36)} & \textbf{19.85 (17.39)} & \textbf{18.05 (15.38)} \\ \hline
\end{tabular}
\end{table*}

\subsection{Individual Task Performance}
\label{sec:task_performance_individual}

Individual performance for each task was recorded throughout the experiment, as shown in Table \ref{tab:nasamatb_online}. Tracking task performance, measured by RMSE in pixels, shows that the RL agent achieved the lowest error at $24.19$ (SD $2.15$), maintaining consistency across all workloads. The RB agent performed well under Underload ($18.48$ (SD $3.79$)) but deteriorated with higher demands. The RLFT agent matched the RL agent under Overload but varied more under lighter conditions (SD $=7.21$). Overall, the RL agent was the most consistent, adapting well across workloads; statistical tests confirmed its superiority over both RB ($U=58.5$, $p<0.05$) and RLFT ($U=64.0$, $p<0.05$).

System monitoring, assessed via response time and failure rate, again favored the RL agent. It had the fastest response time overall ($9.12$ (SD $4.10$) seconds) and maintained speed across all loads. Mann-Whitney U tests showed that the agent significantly outpaced RB ($U=50.0$, $p<0.005$) and RLFT ($U=31.0$, $p<0.001$). The RB agent responded quickly under Underload ($8.62$ (SD $23.32$)) but slowed with added load. RLFT had the slowest and most inconsistent response times, especially under Normal and Overload conditions. For failure rates, the RL agent remained reliable, with no failures in Underload and relatively low rates in heavier conditions. Its performance was significantly better than RLFT ($U=64.0$, $p<0.05$), though not significantly better than RB ($U=126.5$, $p<0.05$). The RB agent’s failure rate rose with workload, while RLFT exhibited the highest failure rate overall, worsening significantly under Overload. When taking response time and failure rate together, the RL agent delivered the most reliable performance across conditions. The RB agent was effective in light workloads but declined under stress. RLFT showed some promise but struggled to scale, particularly as the workload increased.

Resource management favored the RB agent, which had the lowest failure rate across conditions ($5.41$\% (SD $5.15$)). The RL agent’s failure rate increased with workload, from $9.08$\% (SD $11.94$) in Underload to $12.93$\% (SD $18.36$) in Overload. RLFT performed slightly better than RL in Underload but worse in higher loads, peaking at $13.66$\% (SD $19.93$) in Overload. Statistical tests confirmed that the RB agent significantly outperformed both RL ($U=60.0$, $p<0.05$) and RLFT ($U=61.5$, $p<0.05$), highlighting the strength of rule-based control in resource management tasks.

In the communications task, response time and failure rate were assessed across Normal Load and Overload conditions since there are no audio command events during the Underload condition. The RLFT agent had the fastest overall response time at $13.44$ seconds (SD $3.20$), outperforming both RB and RL in Normal ($10.66$ versus $10.96$ seconds) and Overload ($13.90$ versus $15.45$ seconds) conditions. These differences were statistically significant against RB ($U=60.0$, $p<0.05$) and RL ($U=49.0$, $p<0.005$). RL performed well in Normal Load but slowed under Overload, while RB was consistently the slowest ($14.57$ seconds overall), with greater decline under higher load. The RLFT agent also had the lowest overall failure rate at $18.05$\% (SD $15.38$), performing best under Normal Load ($7.27$\% (SD $9.36$)) and remaining lower than others even in Overload ($19.85$\% (SD $17.39$)). These differences were statistically significant compared to RL ($U=64.5$, $p<0.05$) and RB ($U=69.0$, $p<0.05$). RB’s failure rate rose sharply under Overload ($25.92$\% (SD $16.95$)), yielding an overall rate of $24.40$\%. RL started moderately under Normal Load ($12.12$\%) but worsened in Overload ($27.78$\%), resulting in the highest overall failure rate ($25.54$\%). Overall, RLFT was the most effective in the communications task, with the best response time and failure rate across conditions. RB was more stable than RL but less efficient. RL struggled under high load, showing the steepest performance drop; statistical results confirm RLFT’s advantage in terms of both speed and reliability.

\subsection{Task Automation Action Frequency}
\label{sec:results_task_automation}

\begin{figure*}[h!]
\centering
\includegraphics[width=0.7\linewidth]{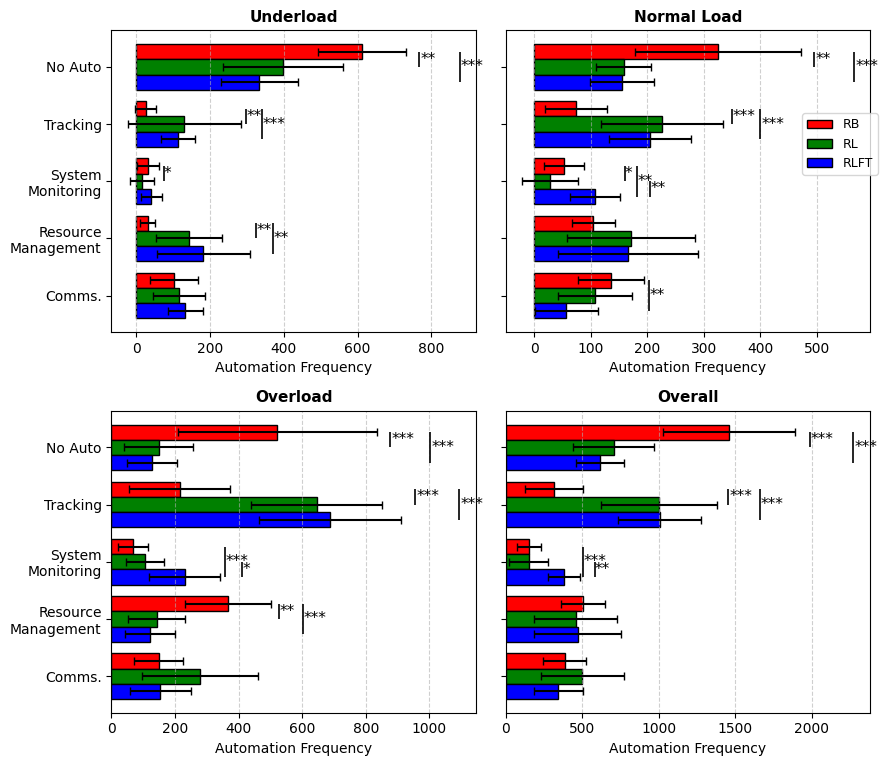}
\caption{Overall task automation action frequency in the NASA Multi-Attribute Task Battery II (MATB-II). Asterisks indicate statistically significant differences between agent conditions -- $p < 0.05$ (*), $p < 0.01$ (**), $p < 0.001$ (***) -- based on Mann–Whitney U tests.}
\label{fig:action_freq}
\end{figure*}

Figure \ref{fig:action_freq} shows automation frequency across the MATB-II tasks as well as the ``No Automation'' case, offering a critical lens through which to interpret performance patterns. A clear relationship emerges between the frequency of task automation and performance effectiveness, revealing distinct strategies and trade-offs among the agents.

In the tracking task, RL and RLFT agents automated the most frequently, while RB did so significantly less. Mann-Whitney U tests confirmed that these differences across all workloads (e.g., Underload: RL versus RB: $U = 19.0$, $p < 0.001$; RLFT versus RB: $U = 11.0$, $p < 0.0001$; Overload: RL versus RB: $U = 5.0$, $p < 0.001$). This frequent automation allowed the RL and RLFT agents to maintain lower RMSE values under high load, especially as tracking requires continuous attention. RL’s consistent performance (RMSE $= 24.10$) reflected adaptive automation, offloading user effort under stress. RB's lower RMSE in Underload ($18.48$) suggests its manual approach worked only when cognitive demands were minimal.

For system monitoring, RL achieved the best results—lowest failure rate ($15.56$\%) and fastest response time ($9.12$ seconds), despite automating the task least. This suggests the RL agent's success stemmed from selective, well-timed interventions. In contrast, the RLFT automated system monitoring the most often but underperformed, highlighting that effectiveness depends not just on automation frequency but also on strategic timing. RB, which matched RL in terms of automation rate, exhibited moderate performance. RL outperformed RB significantly under the Underload ($U = 130.0$, $p = 0.027$) and Normal Load conditions ($U = 131.0$, $p = 0.026$), further supporting the value of targeted automation.

Resource management, a long-term control task, favored the RB agent, which maintained high automation frequency and achieved the lowest failure rates across all conditions. Statistically, RB outperformed RL and RLFT under Overload (RB versus RL: $U = 150.0$, $p = 0.0016$; RB versus RLFT: $U = 172.0$, $p = 0.0004$) and Underload (RB versus RL: $U = 32.5$, $p = 0.0096$; RB versus RLFT: $U = 28.0$, $p = 0.0022$). RL and RLFT likely de-prioritized this task under load, resulting in missed or delayed interventions. This suggests that adaptive agents may struggle in tasks that require persistent, proactive control when faced with competing demands.

\begin{figure}[h!]
\centering
\includegraphics[width=0.9\linewidth]{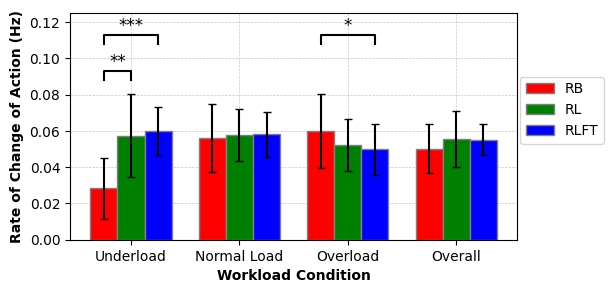}
\caption{Rate of change of actions in the NASA Multi-Attribute Task Battery II (MATB-II). This metric reflects how frequently the automation system switched between tasks, indicating the adaptability and responsiveness of the agent’s control strategy. Asterisks indicate statistically significant differences between agent conditions -- $p < 0.05$ (*), $p < 0.01$ (**), $p < 0.001$ (***) -- based on Mann–Whitney U tests.}
\label{fig:rocoa}
\end{figure}

The RLFT agent was the top performer in the communications task, with the lowest failure rate ($18.05$\%) and the fastest response time ($13.44$ seconds). Notably, it automated the task the least and showed the most stable automation pattern (see Figure \ref{fig:rocoa}), suggesting a selective strategy that supported user engagement without over-automation. In contrast, the RL agent automated the communications task most frequently but had the highest failure rate ($25.54$\%), particularly under the Overload condition, implying that excessive automation may hinder user responsiveness in tasks that require direct input, e.g., interpreting audio commands.

Compared to the broadly trained RL agent, which performed reliably across tasks, RLFT resulted in greater variability. Its personalization benefited the communications task but proved less effective under cognitive overload in tasks such as resource management, where sustained or anticipatory control is critical. This reflects a trade-off in real-time fine-tuning: although personalization can improve task alignment and fluency, it may also reduce robustness and furthermore introduce inconsistency.

\subsection{In Situ Trust, Workload, and Fluency}
In situ subjective ratings of trust, workload, and fluency provide insight into how participants perceived each agent’s automation strategy. After each workload condition, participants rated their experience on a $1$–$5$ Likert scale in response to: 
\textbf{(i)} \textit{``Please rate the level of trust you have in your robot teammate.''}; 
\textbf{(ii)} \textit{``How would you rate the level of mental and physical effort required to complete the task?''}; and, 
\textbf{(iii)} \textit{``How smoothly and efficiently did you and your robot teammate work together during the task?''}. 
These assessments reflect the user experience and complement the objective performance metrics.

\begin{figure}[h!]
\centering
\includegraphics[width=0.9\linewidth]{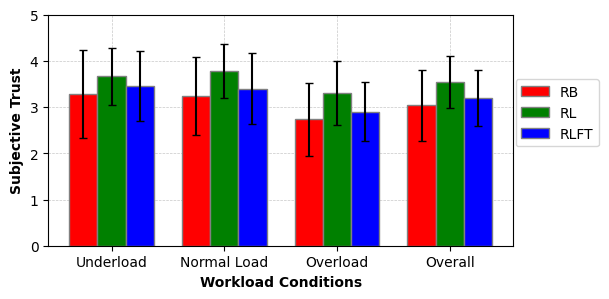}
\caption{Subjective In-Situ Trust in the NASA Multi-Attribute Task Battery II (MATB-II). Asterisks indicate statistically significant differences between agent conditions -- $p < 0.05$ (*), $p < 0.01$ (**), $p < 0.001$ (***) -- based on Mann–Whitney U tests.}
\label{fig:sub_trust}
\end{figure}

Figure \ref{fig:sub_trust} shows that the RL agent received the highest trust ratings across all workload conditions (mean of $3.54$), outperforming RLFT ($3.20$) and RB ($3.04$). Trust peaked under the Normal Load condition, especially for RL ($3.78$), but declined for all agents under Overload. This suggests that limiting automation to one task at a time may not sufficiently reduce cognitive strain. RL's higher trust ratings likely stems from its adaptive usage of physiological data, whereas RLFT showed no added trust benefit over standard RL. Future work could explore transparent or explainable AI tools to further enhance trust in these contexts.

\begin{figure}[h!]
\centering
\includegraphics[width=0.9\linewidth]{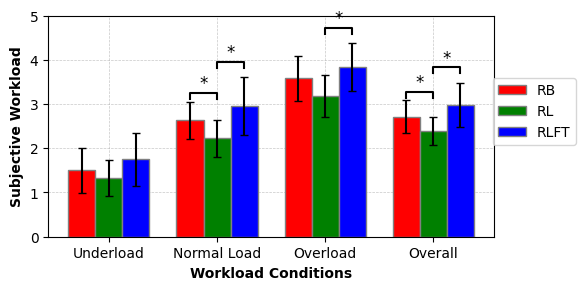}
\caption{Subjective In-Situ Workload in the NASA Multi-Attribute Task Battery II (MATB-II). Asterisks indicate statistically significant differences between agent conditions -- $p < 0.05$ (*), $p < 0.01$ (**), $p < 0.001$ (***) -- based on Mann–Whitney U tests.}
\label{fig:sub_workload}
\end{figure}

Workload ratings, presented in Figure \ref{fig:sub_workload}, show that the RL agent resulted in the lowest perceived workload overall ($2.38$), followed by RB ($2.71$) and RLFT ($2.99$). RL was most effective in reducing effort under the Underload condition ($1.33$) and Normal Load ($2.22$). Although workload increased for all agents under Overload, RL still scored the lowest ($3.19$) and RLFT scored the highest ($3.83$). These results suggest that RL’s workload-aware automation eased cognitive demands, especially in low-stress conditions; RLFT, on the other hand, may have added cognitive strain under pressure, reducing user comfort in high-demand scenarios.

\begin{figure}[h!]
\centering
\includegraphics[width=0.9\linewidth]{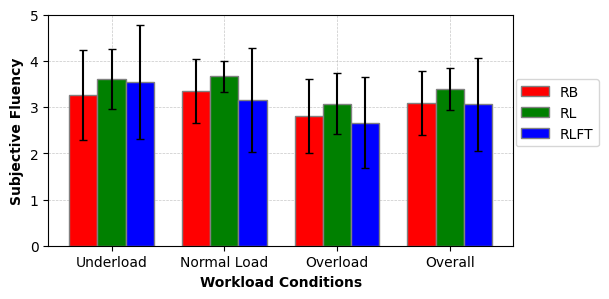}
\caption{Subjective In-Situ Fluency in the NASA Multi-Attribute Task Battery II (MATB-II). Asterisks indicate statistically significant differences between agent conditions -- $p < 0.05$ (*), $p < 0.01$ (**), $p < 0.001$ (***) -- based on Mann–Whitney U tests.}
\label{fig:sub_fluency}
\end{figure}

Fluency ratings in Figure \ref{fig:sub_fluency} showed the RL agent leading overall ($3.40$), followed by RB ($3.09$) and RLFT ($3.06$). Fluency peaked under Normal Load for RL ($3.67$), reflecting the strongest user-agent synchronization at moderate workload. All agents saw fluency decline during Overload, with RL ($3.07$) still outperforming RB ($2.81$) and RLFT ($2.67$). These results suggest that RL’s automation, based on learned user patterns, enhances teamwork perception. In contrast, RLFT’s lower fluency under Overload may stem from real-time adaptations disrupting task flow or timing consistency. This highlights a challenge for fine-tuning -- the problem of balancing personalization with predictability in order to maintain smooth human-robot interaction.

Across trust, workload, and fluency, the RL agent consistently outperformed both RB and RLFT, demonstrating the advantages of RL-driven adaptive automation. The RLFT agent showed no clear benefit over RL, especially under high workload conditions, suggesting that real-time fine-tuning can introduce unpredictability and cognitive strain if not carefully managed. While RLFT performed well in specific tasks like the communications task, its results were inconsistent in complex multitasking scenarios, with higher failure rates and delayed responses under stress. These findings highlight that personalization through fine-tuning must seek to balance adaptability with stability and predictability. Future work should focus on optimizing automation timing, enhancing explainability, and incorporating user feedback to refine human-aware automation without compromising performance or the user experience.

\subsection{Trust, Workload, and Fluency via Standard Questionnaires}
\label{sec:results_questionaire}

The post-session ratings gathered through standardized questionnaires provided additional insight into participants' experiences with the automation systems across the NASA Multi-Attribute Task Battery II (MATB-II). These ratings were based on trust, workload, and fluency, assessed using established scales within the Trust in Automation questionnaire \cite{korber2019theoretical}, the NASA-TLX Workload questionnaire \cite{hart1988development}, and the Fluency questionnaire \cite{hoffman2019evaluating}.

\begin{table*}[h!]
\centering
\caption{Post-trial trust in the NASA Multi-Attribute Task Battery II (MATB-II) via $1$-$5$ Likert scale Trust in Automation questionnaire \cite{korber2019theoretical}.}
\label{tab:post_trust}
\begin{tabular}{|c|l|c|c|c|}
\hline
\textbf{Scale} & \multicolumn{1}{c|}{\textbf{Item}} & \textbf{RB} & \textbf{RL} & \textbf{RLFT} \\ \hline
\multirow{6}{*}{\begin{tabular}[c]{@{}c@{}}Reliability/\\ Competence\end{tabular}} & \begin{tabular}[c]{@{}l@{}}The AI is capable of interpreting \\ situations correctly.\end{tabular} & 3.36 (1.01) & 3.11 (1.16) & \textbf{3.8 (0.78)} \\ \cline{2-5} 
 & The AI works reliably. & 3.52 (0.84) & 3.11 (0.92) & \textbf{3.9 (0.56)} \\ \cline{2-5} 
 & An AI malfunction is likely. & 3.15 (1.16) & \textbf{2.33 (1.0)} & 3.0 (0.81) \\ \cline{2-5} 
 & \begin{tabular}[c]{@{}l@{}}The AI is capable of taking over\\ complicated tasks.\end{tabular} & 3.89 (0.80) & 4.11 (0.60) & \textbf{4.5 (0.52)} \\ \cline{2-5} 
 & The AI might make sporadic errors. & \textbf{3.10 (0.99)} & 3.22 (1.09) & 3.3 (0.94) \\ \cline{2-5} 
 & \begin{tabular}[c]{@{}l@{}}I am confident about the AI’s\\ capabilities.\end{tabular} & 3.57 (0.69) & 3.66 (0.5) & \textbf{4.0 (0.81)} \\ \hline
\multirow{4}{*}{\begin{tabular}[c]{@{}c@{}}Understanding/\\ Predictability\end{tabular}} & The AI state was always clear to me. & 2.94 (1.26) & 2.66 (1.22) & \textbf{3.1 (1.52)} \\ \cline{2-5} 
 & The AI reacts unpredictably. & 3.68 (0.15) & \textbf{3.22 (0.97)} & 3.5 (1.08) \\ \cline{2-5} 
 & \begin{tabular}[c]{@{}l@{}}I was able to understand why things\\ happened.\end{tabular} & 3.52 (0.96) & \textbf{3.77 (0.83)} & 3.7 (1.25) \\ \cline{2-5} 
 & \begin{tabular}[c]{@{}l@{}}It’s difficult to identify what the AI\\ will do next.\end{tabular} & 4.31 (0.94) & 4.44 (1.01) & \textbf{3.9 (1.10)} \\ \hline
\multirow{3}{*}{\begin{tabular}[c]{@{}c@{}}Propensity\\ to Trust\end{tabular}} & \begin{tabular}[c]{@{}l@{}}One should be careful with unfamiliar\\ automated systems (AI).\end{tabular} & \textbf{4.15 (0.89)} & 4.33 (0.70) & 4.4 (0.69) \\ \cline{2-5} 
 & I rather trust an AI than I mistrust it. & 3.42 (0.96) & 3.55 (0.88) & \textbf{3.8 (1.13)} \\ \cline{2-5} 
 & \begin{tabular}[c]{@{}l@{}}Automated systems (AI's) generally\\ work well.\end{tabular} & 3.73 (0.65) & 3.66 (0.70) & \textbf{4.0 (0.66)} \\ \hline
\multirow{2}{*}{\begin{tabular}[c]{@{}c@{}}Trust in\\ Automation\end{tabular}} & I trust the AI. & \textbf{3.42 (0.83)} & 3.33 (0.86) & 3.4 (0.96) \\ \cline{2-5} 
 & I can rely on the AI. & 3.42 (1.07) & 3.66 (0.86) & \textbf{4.0 (0.66)} \\ \hline
\end{tabular}%
\end{table*}

Table \ref{tab:post_trust} shows participants' post-trial trust ratings for the RB, RL, and RLFT agents across four categories: `Reliability/Competence', `Understanding/Predictability', `Propensity to Trust', and `Trust in Automation', using a $1$–$5$ Likert scale. Interpretations are adjusted based on whether higher or lower scores denote greater trust. RLFT generally received the highest trust, especially in Reliability/Competence; here, participants rated its handling of complex tasks highest ($4.5$) and showed more confidence ($4.0$). Negative statements like ``An artificial intelligence (AI) malfunction is likely'' scored lowest under RL and RLFT, indicating greater trust. Understanding/Predictability improved modestly with RLFT, with clearer AI state perception ($3.1$) and better predictability (lower reverse-coded scores). Participants remained cautious about unfamiliar systems across all agents ($>4.0$ in Propensity to Trust), but RLFT scored higher on positive trust statements. In Trust in Automation, RLFT led with the highest ratings for ``I can rely on the AI'' ($4.0$).

Overall, RLFT boosted perceived reliability, competence, and predictability. Trust gains were especially notable in decreased agreement with negatively framed items, suggesting improved behavior and confidence. However, Understanding/Predictability remained a challenge—AI transparency was limited ($3.1$) and behavior predictability was still difficult ($3.9$). High caution scores reflect a baseline skepticism toward AI, indicating performance alone cannot eliminate distrust. Nevertheless, RLFT’s positive impact suggests that fine-tuning does enhance trust, highlighting the need for transparency in order to further improve user confidence. These results emphasize the value of assessing both trust and skepticism, supporting RL with user-aligned fine-tuning as a path to foster greater trust.

\begin{table}[h!]
\centering
\caption{Post-trial subjective workload in the NASA Multi-Attribute Task Battery II (MATB-II) via NASA-TLX questionnaire \cite{hart1988development}.}
\label{tab:post_workload}
\begin{tabular}{|c|c|c|c|}
\hline
\multicolumn{1}{|c|}{\textbf{Item}} & \multicolumn{1}{c|}{\textbf{RB}} & \multicolumn{1}{c|}{\textbf{RL}} & \multicolumn{1}{c|}{\textbf{RLFT}} \\ \hline
\multicolumn{1}{|c|}{Overall Workload} & \multicolumn{1}{c|}{61.4 (13.3)} & \multicolumn{1}{c|}{\textbf{55.8 (14.0)}} & \multicolumn{1}{c|}{65.2 (9.0)} \\ \hline
\multicolumn{1}{|c|}{Mental Demand} & \multicolumn{1}{c|}{78.94 (15.2)} & \multicolumn{1}{c|}{\textbf{69.4 (16.2)}} & \multicolumn{1}{c|}{76.5 (22.6)} \\ \hline
\multicolumn{1}{|c|}{Physical Demand} & \multicolumn{1}{c|}{39.7 (28.9)} & \multicolumn{1}{c|}{\textbf{28.3 (29.5)}} & \multicolumn{1}{c|}{54.5 (24.9)} \\ \hline
Temporal Demand & 66.3 (24.7) & \textbf{61.6 (19.0)} & 69.0 (20.2) \\ \hline
Performance & 25.5 (18.7) & \textbf{34.4 (26.1)} & 30.0 (19.5) \\ \hline
Effort & 65.5 (15.5) & \textbf{57.7 (18.2)} & 74.5 (18.3) \\ \hline
Frustration & 48.9 (28.2) & \textbf{39.4 (26.8)} & 48.0 (27.9) \\ \hline
\end{tabular}
\end{table}

Table \ref{tab:post_workload} summarizes participants' subjective workload ratings using the NASA-TLX questionnaire. Overall workload was lowest under RL ($55.8$), higher under RB ($61.4$), and the highest with RLFT ($65.2$). Mental demand was greatest in RB ($78.9$), followed by RLFT ($76.5$), and then finally the lowest for RL ($69.4$). RL also had the lowest ratings for physical ($28.3$) and temporal demand ($61.6$), while RLFT showed the highest in both ($54.5$ and $69.0$, respectively). Performance ratings (lower is better) slightly worsened from RB ($25.5$) to RL ($34.4$), with RLFT in between ($30.0$). Effort was highest in RLFT ($74.5$) and lowest in RL ($57.7$). Frustration was also lowest under RL ($39.4$) and similarly higher in RB ($48.9$) and RLFT ($48.0$).

Despite RLFT yielding the highest trust ratings, it was found to produce the greatest workload, possibly due to increased cognitive and physical engagement with its more autonomous behavior. Users may have needed to monitor and interpret more complex system actions, raising task demands. Conversely, RL had the lowest workload but also the lowest trust. Its limited interventions likely reduced user burden but offered less support, requiring participants to rely more on manual control. RB consistently fell between RL and RLFT, reflecting moderate demands and predictability that are typical of a rule-based system. Interestingly, subjective performance ratings declined slightly in RL compared to RB and only modestly improved in the case of RLFT. This suggests a mismatch between perceived effectiveness and the value of AI assistance, especially in more dynamic conditions. Overall, these results highlight a tradeoff: more capable and trusted AI (such as RLFT) may increase user workload whereas simpler systems, in contrast, reduce strain but offer less confidence. Thus, effective AI design must balance autonomy, trust, and user effort.

\begin{table}[h!]
\centering
\caption{Post-trial fluency workload in the NASA Multi-Attribute Task Battery II (MATB-II) via $1$-$7$ Likert scale fluency questionnaire \cite{hoffman2019evaluating}.}
\label{tab:post_fluency}
\resizebox{\linewidth}{!}{%
\begin{tabular}{|l|c|c|c|}
\hline
\multicolumn{1}{|c|}{\textbf{Item}} & \textbf{RB} & \textbf{RL} & \textbf{RLFT} \\ \hline
\begin{tabular}[c]{@{}l@{}}The human-robot team worked fluently\\ together.\end{tabular} & 4.68 (1.49) & 4.55 (1.66) & \textbf{5.2 (1.81)} \\ \hline
\begin{tabular}[c]{@{}l@{}}The human was the most important\\ member of the team.\end{tabular} & 6.15 (0.89) & 6.22 (0.66) & \textbf{6.3 (1.05)} \\ \hline
The robot was unintelligent. & 3.31 (1.45) & 3.22 (0.97) & \textbf{2.7 (1.05)} \\ \hline
The robot was trustworthy. & 5.05 (1.12) & 4.66 (1.32) & \textbf{5.1 (0.73)} \\ \hline
The robot was uncooperative. & \textbf{3.05 (1.35)} & 3.55 (1.01) & 3.5 (1.58) \\ \hline
\begin{tabular}[c]{@{}l@{}}The robot contributed to the fluency of\\ the collaboration.\end{tabular} & 4.94 (1.71) & 5.11 (1.61) & \textbf{5.6 (1.77)} \\ \hline
\begin{tabular}[c]{@{}l@{}}The robot was committed to the success\\ of the team.\end{tabular} & 5.15 (1.42) & 5.0 (1.80) & \textbf{5.7 (1.25)} \\ \hline
\begin{tabular}[c]{@{}l@{}}The robot had an important contribution\\ to the success of the team.\end{tabular} & 4.84 (1.34) & 5.22 (1.85) & \textbf{5.4 (1.64)} \\ \hline
\end{tabular}%
}
\end{table}

Table \ref{tab:post_fluency} summarizes the participants' perceived collaboration fluency with AI across the RB, RL, and RLFT conditions, using a $1$–$7$ Likert scale. RLFT yielded the highest overall fluency rating ($5.2$), followed by RB ($4.68$), and then RL ($4.55$). Across all conditions, participants consistently rated the human as the most important team member, especially in the case of RLFT ($6.3$), reinforcing a strong sense of user agency.

Negative items, e.g., ``The robot was unintelligent'' or ``uncooperative'', showed the most favorable RLFT ratings, with the lowest perceived unintelligence ($2.7$) and low uncooperativeness ($3.5$). RL and RB scored slightly worse, suggesting some friction in less fine-tuned agents. RLFT also scored the highest in trustworthiness ($5.1$), contribution to fluency ($5.6$), team commitment ($5.7$), and importance to team performance ($5.4$), reflecting strong integration into the collaborative task.

Participants perceived the most fluent collaboration in the RLFT condition, suggesting that enhanced adaptability and fine-tuning can significantly improve team dynamics. High ratings for trustworthiness, team contribution, and commitment indicate that RLFT was viewed as a supportive, integrated partner. However, while RLFT excelled in tasks like the communication task, it did not consistently outperform RL in more complex, multitasking scenarios. Increased failure rates and slower response times in some cases suggest that real-time fine-tuning can introduce unpredictability, especially under high cognitive load. This underscores the need to balance personalization with system stability and predictability.

Lower scores on negatively worded items, e.g., ``The robot was unintelligent'', further confirm that improved AI competence reduced friction and further enhanced effectiveness. Participants consistently viewed the human as the primary team member, but also recognized the AI system’s growing utility as its capabilities improved; this indicated a balance of control and trust. Slightly higher uncooperativeness ratings in RL and RLFT hint at occasional misalignment between user expectations and autonomous behavior; however, these did not significantly affect overall fluency perceptions in RLFT.

Ultimately, these findings highlight that technical advances must be matched by seamless, user-centered collaboration. While RLFT showed clear potential, personalization alone is not sufficient for complex, high-demand settings. In order to enhance its real-world effectiveness, future work should:
\begin{itemize}
    \item Optimize fine-tuning mechanisms to reduce disruptions, particularly under high cognitive load; 
    \item Investigate adaptive personalization strategies that maintain stability and predictability in task execution; and, 
    \item Incorporate explainable AI methodology to enhance user trust, transparency, and understanding of how the fine-tuned system makes its decisions.
\end{itemize} 
These improvements could ultimately make RLFT more consistently effective in dynamic, real-time environments, leading to smoother and more personalized human-AI interaction.

\section{Conclusion}
\label{sec:conclusion}

This article presents a novel offline reinforcement learning (RL) framework that combines off-policy evaluation (OPE) with human physiological data in order to enable adaptive, responsive human-robot teaming. By using logged interaction data to optimize state representations and reward functions, the proposed method eliminates the need for costly online experimentation, making it suitable for real-world and safety-critical environments. Experimental results from both simulated and human-subject studies demonstrate the effectiveness of OPE in guiding RL design decisions, improving policy performance.

Beyond scalable RL formulation, this work explores bootstrapped human-aware RL, integrating physiological signals into the agent’s state to enable more personalized, context-sensitive behavior. Offline RL agents with human state awareness were shown to consistently outperform rule-based and fine-tuned alternatives in trust, workload, and fluency. However, our findings highlight a tradeoff between personalization and predictability given that fine-tuned agents showed promise but added higher cognitive demands as well as occasional instability.

These results emphasize the importance of balancing autonomy with transparency and user alignment in human-robot collaboration. Future work should focus on refining fine-tuning mechanisms in order to improve stability, developing adaptive strategies that maintain trust under cognitive load, and incorporating explainable AI mechanisms to enhance user understanding and system predictability. Taken wholistically, our proposed framework offers a promising foundation for deploying personalized RL agents that are both effective and user-centered for dynamic, real-world problem environments.



\bibliography{bibliography}
\bibliographystyle{IEEEtran}

\newpage

\vfill

\end{document}